# Feature Selection for Bayesian Evaluation of Trauma Death Risk

L. Jakaite and V. Schetinin

University of Bedfordshire/Department of Computing and Information Systems, Luton, UK

*Abstract*— In the last year more than 70,000 people have been brought to the UK hospitals with serious injuries. Each time a clinician has to urgently take a patient through a screening procedure to make a reliable decision on the trauma treatment. Typically, such procedure comprises around 20 tests; however the condition of a trauma patient remains very difficult to be tested properly. What happens if these tests are ambiguously interpreted, and information about the severity of the injury will come misleading? The mistake in a decision can be fatal – using a mild treatment can put a patient at risk of dying from posttraumatic shock, while using an over-treatment can also cause death. How can we reduce the risk of the death caused by unreliable decisions? It has been shown that probabilistic reasoning, based on the Bayesian methodology of averaging over decision models, allows clinicians to evaluate the uncertainty in decision making. Based on this methodology, in this paper we aim at selecting the most important screening tests, keeping a high performance. We assume that the probabilistic reasoning within the Bayesian methodology allows us to discover new relationships between the screening tests and uncertainty in decisions. In practice, selection of the most informative tests can also reduce the cost of a screening procedure in trauma care centers. In our experiments we use the UK Trauma data to compare the efficiency of the proposed technique in terms of the performance. We also compare the uncertainty in decisions in terms of entropy.

## I. Introduction

As it has been reported in [1], more than 70,000 people have been admitted into the UK hospitals with serious injuries. To make a reliable decision on the trauma treatment, a clinician has to urgently take a patient through a screening procedure which typically comprises around 20 tests [2]. However, the condition of a trauma patient is still very difficult to be tested properly. If the screening tests are ambiguously interpreted, and information about the severity of the injury is misleading, the mistake in a decision can be fatal; the choice of a mild treatment can put a patient at risk of dying from posttraumatic shock, while the choice of an overtreatment can also cause death [1]. How can we reduce the risk of the death caused by unreliable decisions?

It has been shown in [3 - 6] that probabilistic reasoning, based on the Bayesian methodology of averaging over decision models, enables to evaluate the uncertainty in decision making. The use of the Bayesian Model Averaging (BMA) over Decision Trees (DTs) makes decision models interpretable for clinicians as shown in [7]. The Bayesian averaging over DTs (BDTs) enables to select attributes which make the most significant contribution to decisions. Within the Bayesian DTs averaging we would expect that discarding weakest attributes would improve the performance. However, in our experiments, we observed that the performance decreased. Obviously, we can explain that this happened because the discarded attribute was still important for a small amount of the data. If this is the case, then we can expect that the replacement of this attribute by noise will further decrease the performance. Alternatively, we can assume that the weakest attribute makes a contribution to the BMA. Would it be possible to discard the weakest attribute without decreasing the performance? If so, then we can reduce the number of screening tests required for making reliable decisions within BDT methodology.

In theory, BMA methodology is immune to overfitting problem [3]. However, in some cases, overfitting was shown to affect the BMA performance [8]. Based on these results we can assume that if the replacement of the weakest attribute by noise does not decrease the BMA performance, this attribute, making negligible contribution, provides better conditions for mitigating BMA overfitting.

Based on these assumptions, in this paper we aim at selecting the most important screening tests, keeping the BDT performance high. This is important because selection of the most informative screening tests can reduce the cost of a screening procedure in trauma care centers. In our experiments we use the UK Trauma data to compare the efficiency of the proposed BMA technique in terms of the performance. We also compare the uncertainty in decisions in terms of entropy.

Section 2 of the paper describes the bases of BMA over DTs, and section 3 describes the Trauma data used for the experiments. The experimental results are presented in section 4, and section 5 concludes the paper.

## II. Methodology of Bayesian Model Averaging

For a DT given with parameters $\theta$, the predictive distribution is written as an integral over the parameters $\theta$

$$p(y|\mathbf{x},\mathbf{D}) = \int_\theta p(y|\mathbf{x},\boldsymbol{\theta},\mathbf{D})p(\boldsymbol{\theta}|\mathbf{D})d\boldsymbol{\theta}$$

where $y$ is the predicted class (1, ..., $C$), $\mathbf{x} = (x_1, ..., x_m)$ is the $m$-dimensional vector of input, and $\mathbf{D}$ are the given training data.

This integral can be analytically calculated only in simple cases, and in practice part of the integrand, which is the posterior density of $\boldsymbol{\theta}$ conditioned on the data $\mathbf{D}$, $p(\boldsymbol{\theta}|\mathbf{D})$, cannot usually be evaluated. However if values $\boldsymbol{\theta}^{(1)}, ..., \boldsymbol{\theta}^{(N)}$ are the samples drawn from the posterior distribution $p(\boldsymbol{\theta}|\mathbf{D})$, we can write

$$p(y|\mathbf{x},\mathbf{D}) \approx \sum_{i=1}^N p(y|\mathbf{x},\boldsymbol{\theta}^{(i)},\mathbf{D})p(\boldsymbol{\theta}^{(i)}|\mathbf{D})$$
$$= \frac{1}{N}\sum_{i=1}^N p(y|\mathbf{x},\boldsymbol{\theta}^{(i)},\mathbf{D})$$

The above integral can be approximated by using Markov Chain Monte Carlo (MCMC) technique [3]. To perform such an approximation, we need to run a Markov Chain until it has converged to a stationary distribution. Then we can collect $N$ random samples from the posterior $p(\boldsymbol{\theta}|\mathbf{D})$ to calculate the desired predictive posterior density.

Using DTs for the classification, we need to find the probability $\varphi_{tj}$ with which an input $\mathbf{x}$ is assigned by terminal node $t = 1, ..., k$ to the $j$th class, where $k$ is the number of terminal nodes in the DT. The DT parameters are defined as $\boldsymbol{\theta} = (s_i^{pos}, s_i^{var}, s_i^{rule})$, $i = 1, ..., k-1$, where $s_i^{pos}$, $s_i^{var}$ and $s_i^{rule}$ define the *position*, *predictor* and *rule* of each splitting node, respectively. For these parameters the priors can be specified as follows. First we can define a maximal number of splitting nodes, say, $s_{max} = n - 1$. Second we draw any of the $m$ predictors from a uniform discrete distribution $U(1, ..., m)$ and assign $s_i^{var} \in \{1, ..., m\}$.

Finally the candidate value for the splitting variable $x_j = s_i^{var}$ can be drawn from a discrete distribution $U(x_j^{(1)}, ..., x_j^{(L)})$, where $L$ is the number of possible splitting rules for variable $x_j$, either categorical or continuous. Such priors allow us to explore DTs which split data in as many ways as possible. However the DTs with different numbers of splitting nodes should be explored in the same proportions [3].

To sample DTs of a variable dimensionality, the MCMC technique exploits the Reversible Jump extension [3]. To implement the RJ MCMC technique, Denison *et al.* [3] and Chipman *et al.* [6] have suggested exploring the posterior probability by using the following types of moves:

1. *Birth*. Randomly split the data points falling in one of the terminal nodes by a new splitting node with the variable and rule drawn from the corresponding priors.
2. *Death*. Randomly pick a splitting node with two terminal nodes and assign it to be one terminal with the united data points.
3. *Change-split*. Randomly pick a splitting node and assign it a new splitting variable and rule drawn from the corresponding priors.
4. *Change-rule*. Randomly pick a splitting node and assign it a new rule drawn from a given prior.

The first two moves, *birth* and *death*, are reversible and change the dimensionality of $\boldsymbol{\theta}$. The remaining moves provide jumps within the current dimensionality of $\boldsymbol{\theta}$. Note that the *change-split* move is included to make "large" jumps which potentially increase the chance of sampling from a maximal posterior whilst the *change-rule* move does "local" jumps.

The RJ MCMC technique starts drawing samples from a DT consisting of one splitting node whose parameters were randomly assigned within the predefined priors. So we need to run the Markov Chain while a DT grows and its likelihood is unstable. This phase is said *burn-in* and it should be preset enough long in order to stabilize the Markov Chain. When the Markov Chain will be enough stable, we can start sampling. This phase is said *post burn-in*.

### III. THE PROPOSED METHOD

To test the assumptions made in section I, we propose two methods – the first is based on selection of DTs ensemble, and the second is based on the randomization of variables. The selection technique aims to omit the DTs which use the weakest variable. The randomization technique aims to provide better conditions for mitigating DT ensemble overfitting. In the following sections we test and compare these techniques on the Trauma data.

### IV. EXPERIMENTS AND RESULTS

#### A. Trauma Data

The Trauma data collected at the Royal London Hospital comprises 16 screening tests and attributes and the outcome (lived or died) for 316 injured patients. Among these variables 5 are continuous and 11 are categorical, see Table 1.

#### B. Variable's Importance

In our experiments we collected 10,000 DTs during post burn-in phase after sampling 200,000 DTs during burn-in phase. The sampling rate for post burn-in phase was 7; the number of minimal data instances allowed in DT nodes was 3; the acceptance rate was around 0.25.

Having obtained the ensemble of DTs, we estimated the importance of all 16 variables for the prediction. The estimates were calculated as the posterior probabilities of variables used in the DTs ensemble as shown in Fig. 1.

Table 1 Screening tests and attributes of the Trauma data

| No | Screening Tests and Attributes | Type |
|---|---|---|
| 1 | Age | Continuous |
| 2 | Gender: Male = 1, Female = 0. | 0,1 |
| 3 | Injury type: Blunt = 1, penetrating = 0 | 0,1 |
| 4 | Head injury, no injury = 0 | 0,1,2,3,4,5,6 |
| 5 | Facial injury | 0,1,2,3,4 |
| 6 | Chest injury | 0,1,2,3,4,5,6 |
| 7 | Abdominal or pelvic contents injury | 0,1,2,3,4,5 |
| 8 | Limbs or bony pelvis injury | 0,1,2,3,4,5 |
| 9 | External injury | 0,1,2,3 |
| 10 | Respiration rate | Continuous |
| 11 | Systolic blood pressure | Continuous |
| 12 | Glasgow coma score (GCS) eye response | 0,1,2,3,4 |
| 13 | GCS motor response | 0,1,2,3,4,5,6 |
| 14 | GCS verbal response | 0,1,2,3,4,5 |
| 15 | Oximetry | Continuous |
| 16 | Heart rate | Continuous |
| 17 | Died = 1, living = 0. | 0,1 |

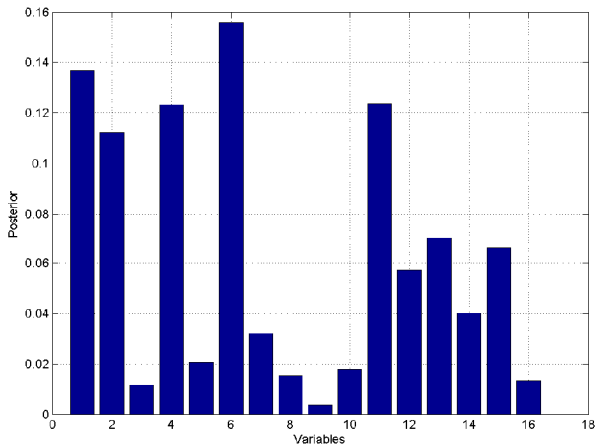

Fig. 1 Posterior probabilities of variables used in the ensemble

From Fig. 1 we can observe that the posterior probability of variable 9 is the smallest, around 0.005, while the maximal value is around 0.16 for variable 6. Therefore we can assume that the variable 9 makes negligible contribution to the ensemble's outcome.

To test our assumptions, we aim to discard this variable from the Trauma data. Table 2 shows the maximal values of loglikelihoods calculated within 5-fold cross-validation for two sets including 16 and 16\9 variables. From this table, we can observe that the loglikelihood value for the 16\9 set becomes greater than that for the set of all 16 variables. However the performance of the ensemble using the set of 16\9 variables is slightly fewer than that using the set of 16 variables. This can happen because the ensemble using the set of 16\9 variables becomes more overfitted to the training data. Thus, we can conclude that the weakest variable 9 provides better conditions for mitigating the DT ensemble overfitting.

Table 2 Maximal loglikelihoods, $L_{16}$ and $L_{16\backslash 9}$, performances and entropies of the ensembles using sets of 16 and 16\9 variables within 5-fold cross-validation

| Fold | Loglikelihood ($L_{16}$) | Loglikelihood ($L_{16\backslash 9}$) |
|---|---|---|
| 1 | -36.14 | -33.4 |
| 2 | -44.29 | -37.33 |
| 3 | -37.23 | -33.19 |
| 4 | -36.24 | -38.37 |
| 5 | -37.98 | -40.41 |
| Loglikelihood | -38.37 ±3.39 | -36.48 ±3.23 |
| Performance, % | 84.66 ±4.70 | 83.85 ±6.95 |
| Entropy | 29.8 ±2.1 | 30.0 ±4.7 |

*C. Selection of DT Ensemble*

As shown above, the presence of the weakest variable has the positive effect on mitigating overfitting of the DT ensemble. This means that the DT ensemble should use all 16 input variables during sampling, but then we can exclude those DTs which use the weakest variable 9. After such selection of DTs there is no need to use the variable 9.

In our experiments this technique was tested within 5-fold cross-validation and results shown in Table 3 which compares the performance of the original DT ensemble using all 16 variables with the performance of the selected ensemble. This table also shows the number of DTs omitted after the selection.

Table 3 Performances and entropies of the original and selected ensembles within 5-fold cross-validation

| | Original ensemble | | | Selected ensemble | |
|---|---|---|---|---|---|
| Fold | Performance, % | Entropy | BDTs omitted | Performance, % | Entropy |
| 1 | 85.93 | 26.47 | 314 | 85.93 | 26.46 |
| 2 | 80.95 | 28.89 | 467 | 80.95 | 28.91 |
| 3 | 84.13 | 31.80 | 217 | 84.13 | 31.79 |
| 4 | 82.54 | 32.05 | 631 | 82.54 | 32.04 |
| 5 | 87.30 | 31.44 | 336 | 87.30 | 31.46 |
| | 84.17±2.54 | 30.13±2.40 | 393±160 | 84.17±2.54 | 30.13±2.40 |

This table show that the performance of the selected ensemble using 16\9 variables is the same as that of the original ensemble using all 16 input variables. The entropies in decisions are also the same. Thus from this experiment we can see that proposed technique allows us to use a reduced set of variables.

*D. Addition of Noise to Variables*

In our experiments some amount of noise added to a weak variable can provide better conditions for mitigating DTs ensemble overfitting. Therefore, we can assume that the addition of noise to all variables will further improve conditions for mitigating DTs ensemble overfitting.

To test this assumption we removed the variable 9 and added a uniform noise to the remaining 15 variables. The intensity of the noise was 0.01. Table 4 shows the performances of DT ensembles using the set of 16 variables and the set of 16\9 with the added noise compared within 5-fold cross-validation. This table shows that the performance of the DT ensemble using the set of 16\9 + noise is better on 2%, than that of the ensemble using the original 16 inputs.

Table 4 Performance and entropy of the ensembles using 16 variables and 16\9 variables with noise estimated within 5-fold cross-validation

| Fold | Trauma (16 variables) | | Trauma (16\9 variables + noise) | |
|---|---|---|---|---|
| | Performance, % | Entropy | Performance, % | Entropy |
| 1 | 84.37 | 26.63 | 85.93 | 26.52 |
| 2 | 79.36 | 28.70 | 80.95 | 30.31 |
| 3 | 84.12 | 32.54 | 88.88 | 32.47 |
| 4 | 88.88 | 31.24 | 90.47 | 32.00 |
| 5 | 88.88 | 28.11 | 88.88 | 32.65 |
| | 85.13± 7.94 | 29.4± 4.80 | 87.03± 7.54 | 30.8± 5.10 |

Thus we can see that the addition of the noise to the set of 16\9 variables allows us to exclude the weakest variable 9 and, at the same time, this enables to improve the DT ensemble performance.

## V. CONCLUSIONS & DISCUSSION

We have expected that discarding weakest attributes can improve the performance of the BDT ensemble. However, in our experiments, the performance has oppositely decreased. We have assumed that this happened because the discarded weakest attribute was still important for a small amount of the data. Alternatively, we have assumed that the weakest attribute makes a noticeable contribution to the BDT ensemble's outcome. The question was would it be possible to discard the weakest attribute without decreasing the performance? This is important for clinical practice if the number of screening tests required for making reliable decisions within BDT methodology can be reduced.

Although BMA methodology in theory is immune to overfitting problem, in some cases, it was shown that overfitting affects the BMA performance. We have observed that the replacement of the weakest attribute by noise did not decrease the BDT performance, and therefore this attribute, making negligible contribution, provided better conditions for mitigating BDT ensemble overfitting.

In this paper we aimed at selecting the most important screening tests, keeping the BDT performance high. In our experiments we used the UK Trauma data to compare the efficiency of the proposed technique in terms of the performance. We also compare the uncertainty in decisions in terms of entropy.

As a result we found that the proposed techniques allow clinicians to reduce number of screening tests, keeping the performance and reliability of making decisions high. The optimized solutions can reduce the cost of a screening procedure in trauma care centers.

## REFERENCES

1. Royal Society for the Prevention of Accidents available at http://www.rospa.com/factsheets
2. Trauma Audit and Research Network available at http://www.tarn.ac.uk
3. Denison D, Holmes C, Malick B, Smith A (2002) Bayesian methods for nonlinear classification and regression. Willey
4. Breiman L, Friedman J, Olshen R, Stone C (1984) Classification and regression trees. Belmont, CA: Wadsworth.
5. Buntine W (1992) Learning classification trees. Statistics and Computing 2: 63-73
6. Chipman H, George E, McCullock R (1998) Bayesian CART model search, J. American Statistics, 93: 935-960
7. Schetinin V et al. (2007) Confident Interpretation of Bayesian Decision Trees for Clinical Applications. IEEE Transaction on Information Technology in Biomedicine, Volume 11, Issue 3, 312-319
8. Domingos P (2000) Bayesian Averaging of Classifiers and the Overfitting Problem, Proc. 17th International Conf. on Machine Learning, San Francisco, CA, 2000, 223-230

Author: L Jakaite  
Institute: University of Bedfordshire  
Street: Park Square  
City: Luton  
Country: UK  
Email: Livija.Jakaite@gmail.com